\documentclass[a4paper,11pt]{article}
\usepackage[hyperref]{acl2021}
\usepackage{times}
\usepackage{latexsym}

\usepackage{microtype}
\usepackage{CJK}
\usepackage{multirow}
\usepackage{tikz}
\usepackage{tikz-qtree}
\usepackage{graphicx}
\usepackage{booktabs}
\usepackage{amsmath}
\usepackage{amssymb}
\usepackage{enumitem}

\aclfinalcopy 
\def\aclpaperid{371} 

\newcommand{\cev}[1]{\reflectbox{\ensuremath{\vec{\reflectbox{\ensuremath{#1}}}}}}

\title{A Top-Down Neural Architecture towards Text-Level Parsing \\of Discourse Rhetorical Structure}

\author{Longyin Zhang$^{1,2}$, Yuqing Xing$^{1,2}$, Fang Kong$^{1,2}$\thanks{~~Corresponding author}, Peifeng Li$^{1,2}$, Guodong Zhou$^{1,2}$ \\
  1. Institute of Artificial Intelligence, Soochow University, China \\
  2. School of Computer Science and Technology, Soochow University, China \\
  \texttt{\{lyzhang9,yqxing\}@stu.suda.edu.cn} \\
  \texttt{\{kongfang,pfli,gdzhou\}@suda.edu.cn} \\}

\date{}

\begin{document}
\begin{CJK}{UTF8}{gkai}
\maketitle
\begin{abstract}
Due to its great importance in deep natural language understanding and various down-stream applications, text-level parsing of discourse rhetorical structure (DRS) has been drawing more and more attention in recent years.
However, all the previous studies on text-level discourse parsing adopt bottom-up approaches, which much limit the DRS determination on local information and fail to well benefit from global information of the overall discourse.
In this paper, we justify from both computational and perceptive points-of-view that the top-down architecture is more suitable for text-level DRS parsing. On the basis, we propose a top-down neural architecture toward text-level DRS parsing.
In particular, we cast discourse parsing as a recursive split point ranking task, where a split point is classified to different levels according to its rank and the elementary discourse units (EDUs) associated with it are arranged accordingly. In this way, we can determine the complete DRS as a hierarchical tree structure via an encoder-decoder with an internal stack. Experimentation on both the English RST-DT corpus and the Chinese CDTB corpus shows the great effectiveness of our proposed top-down approach towards text-level DRS parsing.
\end{abstract}

\section{Introduction}
\begin{figure}[ht]
  \begin{small}
  \begin{center}
    \includegraphics[scale=1]{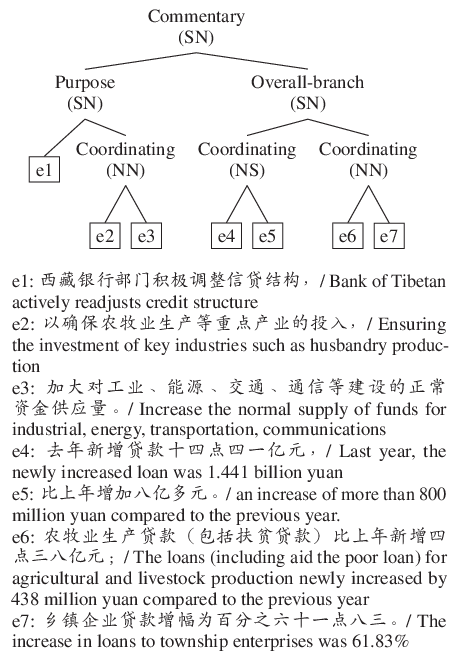}
    \caption{\label{fg_example}An example for DRS parsing, where the text consists of 3 sentences containing 7 EDUs.}
  \end{center}
  \end{small}
\end{figure}

Text-level parsing of discourse rhetorical structure (DRS) aims to identify the overall discourse structure and the rhetorical relations between discourse units in a text.
As a fundamental research topic in natural language processing, text-level DRS parsing plays an important role in text understanding and can benefit various down-stream applications, such as document summarization~\cite{W16-5903}, sentiment analysis~\cite{P16-1032}, text categorization~\cite{ji-smith-2017-neural}, pronoun resolution~\cite{DBLP:conf/nlpcc/ShengKZ17} and event temporal relation identification~\cite{Dia2019}.

According to Rhetorical Structure Theory~(RST)~\cite{RST-1988}, a text can be presented by a hierarchical tree structure known as a Discourse Tree (DT).
Figure~\ref{fg_example} illustrates an excerpt with its gold standard DRS from article chtb\_0005 in the Chinese CDTB (Connective-driven Discourse Treebank) corpus~\cite{D14-1224}. We can find that, in the DT, each leaf node corresponds to an elementary discourse unit (EDU), and various EDUs are recursively combined into high level larger discourse units in a bottom-up fashion. In this example, 7 EDUs are connected by 6 rhetorical relations, while in each non-terminal node, the rhetorical relation and the nuclearity type are labeled.
Correspondingly, text-level DRS parsing consists of three components, i.e., bare DRS generation (hierarchical span determination), rhetorical nuclearity determination and rhetorical relation classification.

During the past decade, text-level DRS parsing has been drawing more and more attention and achieved certain success~\cite{hernault2010hilda,joty2013combining,feng2014linear,ji2014representation,heilman2015fast,li2016discourse,braud2017cross,yu2018transition}.
However, all the previous studies on text-level DRS parsing adopt bottom-up approaches. That is, adjacent EDUs are recursively combined into high-level larger text spans by rhetorical relations to form a final discourse tree in a bottom-up way. In this paper, we justify that compared with a bottom-up approach, a top-down approach may be more suitable for text-level DRS parsing from two points-of-view, 

\begin{itemize}[leftmargin=*]
  \item From the computational view, only local information (i.e., the constructed DRS subtrees and their context) can be naturally employed to determine the upper layer structure in the bottom-up fashion. Due to the overwhelming ambiguities at the discourse level, global information, such as the macro topic or structure of the discourse, should be well exploited to restrict the final DRS, so as to play its important role. From the computational view, a top-down approach can make better use of global information.

  \item From the perceptive view, when people read an article or prepare a manuscript, they normally go from coarse to fine, from general to specific. That is, people tend to first have a general sense of the theme of the article, and then go deep to understand the details. Normally, the organization of the article is much limited by its theme. For text-level DRS parsing, a top-down approach can better grasp the overall DRS of a text and conform to the human perception process.

\end{itemize}

Additionally, just noted as Li et al.~\shortcite{D14-1224}, they employed a top-down strategy in the Chinese CDTB annotation practice. That is, a top-down approach is consistent with the annotation practice of a DRS corpus.
In this paper, we propose a top-down neural architecture to text-level DRS parsing. In particular, we cast top-down text-level DRS parsing as a recursive split point ranking task, where various EDUs associated with split points are arranged in different levels according to the rank of the split point. In this way, we can determine the complete DRS as a hierarchical tree structure via an encoder-decoder with an internal stack. It is worthwhile to mention that, at each time step, we use the Biaffine Attention mechanism~\cite{ICLR-2017} to compute the attention vector and determine the next split point, along with the corresponding nuclearity and relation jointly.

\section{Related Work}

In the literature, previous studies on text-level discourse parsing can be classified into two categories, probabilistic CKY-like approaches~\cite{hernault2010hilda,joty2013combining,feng2014linear,li2014recursive,li2016discourse} and transition-based approaches~\cite{li2014text,ji2014representation,heilman2015fast,wang2017two,braud2017cross,yu2018transition}.

Probabilistic CKY-like approaches normally exploit various kinds of lexical, syntactic and semantic features to compute the probability of the relation between the EDUs, and select the two EDUs with the highest relational probability to merge into one text span. In this way, the final discourse tree is generated. Recently, various deep learning models are employed to capture hidden information to compute the relational probability, e.g. recursive deep models~\cite{li2014recursive}, and attention-based hierarchical neural network models~\cite{li2016discourse}.
As an alternative, transition-based approaches employ the dependency structure to directly represent the relations between EDUs. Li et al.~\shortcite{li2014text} first build a discourse dependency treebank by converting the RST-DT corpus and then apply graph based dependency parsing techniques to discourse parsing. Ji et al.~\shortcite{ji2014representation} propose a shift-reduce discourse parser using a representation learning approach to achieve the state-of-the-art performance.
Wang et al.~\shortcite{wang2017two} propose a pipelined two-stage parsing approach. First, a transition-based model is employed to parse a bare discourse tree. Then, an independent relation labeller is adopted to determine discourse relations.
Braud et al.~\shortcite{braud2017cross} present two variants of transition-based discourse parsing using a feed-forward neural network model.
Yu et al. \shortcite{yu2018transition} build a transition based RST parser with implicit syntactic features. In particular, the information of sentence boundaries and paragraph boundaries is embedded as additional features.

It is worthwhile to emphasize that, all the above studies on text-level discourse parsing employ the bottom-up approaches.
So far, only Lin et al.~\shortcite{lin-etal-2019-unified} and Liu et al.~\shortcite{liu-etal-2019-hierarchical} make the preliminary explorations on constructing sentence-level DTs in a top-down fashion. Lin et al.~\shortcite{lin-etal-2019-unified} proposed a framework for both the EDU segmenter and the sentence-level discourse parser uniformly. Following the work of Lin et al.~\shortcite{lin-etal-2019-unified}, Liu et al.~\shortcite{liu-etal-2019-hierarchical} proposed hierarchical pointer network for better dependency and sentence-level discourse parsing. However, both studies consider merely sentence-level discourse parsing. While it is simple but effective to encode entire sentence sequentially, entire text-level discourse larger than sentence, such as paragraph and document, is obviously much more complicated. Statistics on the RST-DT corpus show each sentence only contains 2.5 EDUs on average while each document contains 55.6 EDUs on average. The representation for large text span can impact the parsing performance very much.

In this paper, we present a top-down neural architecture to text-level DRS parsing.
Different from Lin et al.~\shortcite{lin-etal-2019-unified} and Liu et al.~\shortcite{liu-etal-2019-hierarchical}, we propose a hierarchical discourse encoder to better present the text span using both EDUs and split points. Benefiting from effective representation for large text spans, our text-level discourse parser achieves competitive or even better results than those best reported discourse parsers either neural or non-neural with hand-engineered features.

\section{Top-down Neural Architecture}
Our top-down neural architecture consists of three parts, i.e., EDU Encoder, Split Point Encoder and Attention-based Encoder-Decoder. Among them, the EDU encoder and the split point encoder are responsible for representing the EDUs and the split points, respectively. Different from Lin et al.~\shortcite{lin-etal-2019-unified} and Liu et al.~\shortcite{liu-etal-2019-hierarchical}, we combine the representation of both EDUs and split points hierarchically to better represent the text span rather than only using the representation of the last EDU as the representation of the text span. In this way, the global information can be exploited for our text-level discourse parsing.
In the following, we take Figure~\ref{fg_example} as the example to illustrate the architecture.

\subsection{EDU Encoder}\label{sect_edu_encoder}
\begin{figure}
  \begin{small}
  \begin{center}
    \includegraphics[scale=0.5]{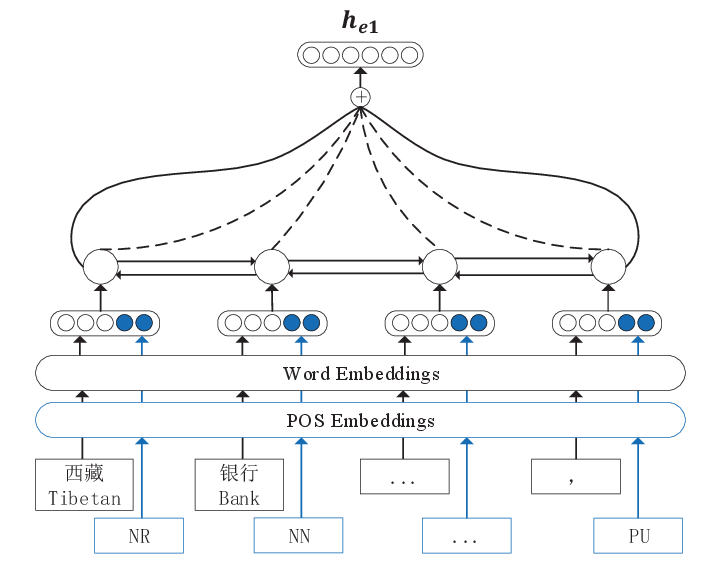}
    \caption{\label{fg_edu_encoder}Architecture of the EDU encoder. }
  \end{center}
  \end{small}
\end{figure}

Figure~\ref{fg_edu_encoder} illustrates the detailed structure of the EDU Encoder.

For a given discourse $D=\{E_1,\dots,E_N\}$, where $N$ means the number of EDUs, $E_k$ is the $k$th EDU.
The EDU encoder is responsible for encoding each EDU.
For $\forall E_k \in D$, $E_k=\{w_{1},w_{2},\dots,w_{n}\}$, where $w_{i}$ means the $i$th word of $E_k$ and $n$ is the number of words, we first concatenate the word embedding and the POS embedding for each word. Then, the combined vectors are fed into the bi-directional GRU network~\cite{cho-etal-2014-learning}. The output of the $i$th word is $h_{i}$, and the last states of BiGRU in both directions are denoted as $h_{\vec{s}}$ and $h_{\cev{s}}$ (i.e., $h_{\vec{s}}=h_{\vec{n}}$, $h_{\cev{s}}=h_{\cev{1}}$).

Considering the different importance of each word in a given EDU, we employ a self-attention mechanism to calculate the weight of each word. Eq.~\ref{eq_edu_weight} shows the weight calculation formula, where we take the dot product of a learnable vector $q$ and $h_{i}$ as the weight of the $i$th word in the EDU.

\begin{equation}\label{eq_edu_weight}
w_{i}=\frac{q^{T} h_{i}}{\sum q^{T} h_{j}}
\end{equation}

In this way, we can achieve the encoding $h_{ek}$ of the $k$th EDU in given discourse $D$.

\begin{equation}\label{eq_edu}
h_{ek}=\left[
          \begin{array}{c}
            h_{\vec{s}} \\
            h_{\cev{s}} \\
          \end{array}
        \right]
        +\sum w_{i} h_{i}
\end{equation}

\subsection{Split Point Encoder}\label{sect_split_encoder}
In this paper, we call the split position between any two EDUs the split point. A discourse containing $n$ EDUs has $n-1$ split points. For example, Figure~\ref{fg_example} contains 7 EDUs and 6 split points. The split point encoder is responsible for encoding each split point. In our model, we use the both EDUs on the left and right sides of the split point to compute the split point representation.

After encoding each EDU using the EDU encoder, we can get the sequence of encoded EDUs $h_e =\{ h_{e1},\dots,h_{eN}\}$, which are further fed into a bi-directional GRU network to get the final sequence of encoded EDUs $h'_e =\{ h'_{e1},\dots,h'_{eN}\}$.

For the convenience of calculation, we first add two additional zero vectors on the start and end of the EDU sequence as stubs.
Then, we use a convolutional network to compute the final split point representation. Here, the width of the convolution kernel is set to 2, and the Rectified Linear Unit ($ReLU$) activation function is employed to map the input $h'_e =\{h'_{e0}, h'_{e1},\dots,h'_{eN},h'_{e(N+1)}\}$ to the output $h_s =\{ h_{s0}, h_{s1},\dots,h_{sN}\}$.

\begin{figure}
  \begin{small}
  \begin{center}
    \includegraphics[scale=0.6]{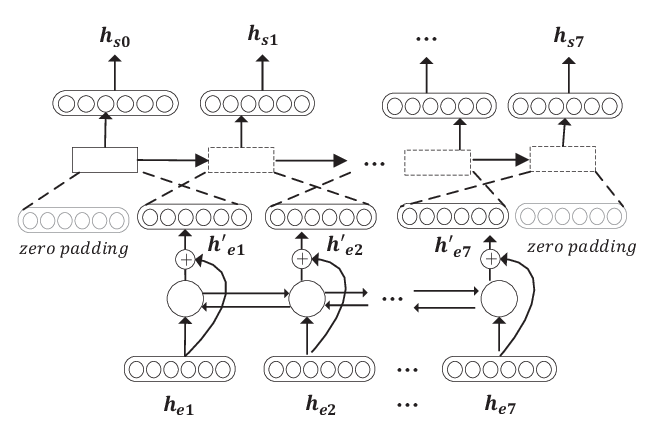}
    \caption{\label{fg_sp_encoder}Architecture of the split point encoder. }
  \end{center}
  \end{small}
\end{figure}

Figure~\ref{fg_sp_encoder} takes the example as shown in Figure~\ref{fg_example} to demonstrate the working procedure of the split point encoder. The input is the achieved 7 EDU encoding results during the EDU encoder stage, i.e., the vector sequence $\{h_{e1}\ldots h_{e7}\}$. The output is the 8 split point representation vectors $\{h_{s0}\ldots h_{s7}\}$, where, the first and last vectors are just stubs and the remaining 6 vectors are meaningful outputs for following stages.

\subsection{Attention-based Encoder-Decoder on Split Point Ranking}\label{sect_en_decoder}
After achieving the representation of each split point, an encoder-decoder with an internal stack is employed to rank the split points and indirectly get the predicted discourse parse tree.

\begin{figure}
  \begin{small}
  \begin{center}
    \includegraphics[scale=0.92]{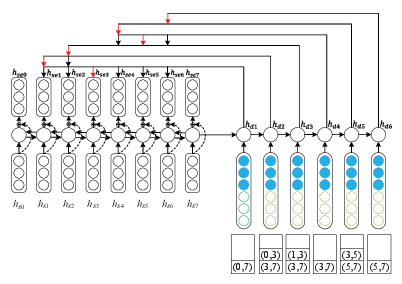}
    \caption{\label{fg_en_decoder}A parsing example of the attention-based encoder-decoder. }
  \end{center}
  \end{small}
\end{figure}

Figure~\ref{fg_en_decoder} shows the complete encoder-decoder framework, where the left part shows the encoder. Here, the achieved split point representation vectors $h_s =\{ h_{s0}, h_{s1},\dots,h_{sN}\}$ are fed into a bi-directional GRU network to get the output $h_{se} =\{ h_{se0}, h_{se1},\dots,h_{seN}\}$.
At the same time, the combination of the last states of the bi-directional GRU network in both directions are taken as the initial state of the decoder.
During the decoder stage, a uni-directional GRU network with an internal stack is employed for our discourse parser. Initially, the stack contains only one element, i.e., the index pair of the first and the last split points of the complete discourse $(0,N)$.
At each decoding step, the index pair of the boundary split points is first popped from the top of the stack. Suppose the index pair is $(l,r)$ at the $j$th step. Then, the encoding output $h_{sel}$ and $h_{ser}$ are concatenated to form the input of the decoder. While the decoder output at the $j$th step represented by $h_{dj}$.
After that, we adopt the Biaffine Attention mechanism to the encoder output corresponding to the split points between the boundary split points (i.e., $h_{sem}$, $\forall m, l \leq m \leq r$) and the decoder output $h_{dj}$. Finally, the split point with the largest score is selected as the final result of this time. If there are still unselected split points for the new text spans formed by this decision, they are pushed onto the stack for following steps.

Figure~\ref{fg_en_decoder} shows the parsing steps of the example shown in Figure~\ref{fg_example}. Here, the arrows in red indicate the selected split points at each time step. $h_{se0}$ and $h_{se7}$ represent the start and end points of the given discourse, and do not participate in the split point selection during decoding. In particular, the stack is first initialized with containing only one element $(0,7)$. That is, all EDUs form a complete text span at the very beginning, and we feed the concatenated vector $[h_{e0};h_{e7}]$ into the decoder to achieve the output $h_{d1}$. Then, the weight is computed using $h_{d1}$ and the results of the encoder corresponding to the 6 split points between the number 0 and the number 7, i.e., $h_{se1}\ldots h_{se6}$. In this example, since the split point 3 has the largest weight, the text span is split into two parts, i.e., $(0,3)$ and $(3,7)$. Because there are still unselected split points in the text span $(0,3)$ and $(3,7)$, we push them onto the stack. In this way, we get one split point at each step. After six iterations, the discourse tree is built.

\subsection{Biaffine Attention on Text-level DRS Parsing}\label{sect_attention}

After achieving the split point representation, we adopt the Biaffine Attention mechanism to determine the split point, nuclearity and discourse relation jointly.
Since applying smaller multi-layer perceptrons (MLPs) to the recurrent output states before the biaffine classifier has the advantage of stripping away information not relevant to the current decision, we first employ a one-layer perceptron to the output vectors of the encoder $h_{sei}$ and the decoder $h_{dj}$ with $ReLU$ as its activation function. The converted vectors are denoted by $h'_{sei}$ and $h'_{dj}$.
Then, we compute the biaffine attention score function. 

\begin{equation}\label{eq_attention_fun}
\begin{split}
s^{i}_{j} = {h'_{sei}}^{T} Wh'_{dj} + Uh'_{sei} + Vh'_{dj}+b; \\
  W \in \mathbb{R}^{m\times k \times n}, U \in \mathbb{R}^{k\times m}, V \in \mathbb{R}^{k\times n}, s^{i}_{j} \in \mathbb{R}^{k}
\end{split}
\end{equation}

\noindent{where $W,U,V,b$ are parameters, denoting the weight matrix of the bi-linear term, the two weight vectors of the linear terms, and the bias vector, respectively, $s^{i}_{j}$ means the score of the $i$th split point over different categories, and the $k$ denotes the number of categories (for split point determination, $k=1$; for nuclearity determination, $k=3$; for discourse relation classification, $k=18$ in English and $k=16$ in Chinese). In this way, we can determine the split point, nuclearity and discourse relation jointly.}

From Eq.~\ref{eq_attention_fun}, we can find that the biaffine attention score function contains three parts, the encoding output, the decoding output, and the combination of the encoder and the decoder in a bilinear way. Among them, the encoding output can be viewed as the information about the current split point, while the decoding output indicates the information about the boundary points and the historical split point.

\subsection{Model Training}
In comparison with transition-based approaches, our approach can not only maintain a linear parsing time, but also perform batch training and decoding in parallel.
In particular, we optimize our discourse parsing model using the Negative Log Likelihood Loss (NLL Loss), which consists of three parts, i.e., the Split Point Prediction Loss ($L_{s}$), the Nuclearity Prediction Loss ($L_{n}$), and the Relation Prediction Loss ($L_{r}$). Among them, the split point prediction loss is used to maximize the probability of selecting the correct split point at each decoding step. Here, we use Eq.~\ref{eq_ls} to compute the loss, assuming that the correct split point number at the $i$th step of the decoder is $j$.

\begin{equation}\label{eq_ls}
L_{s} = \sum_{batch} \sum_{steps} -\log(\hat{p}_{i}^{s}|\theta)
\end{equation}

\begin{equation}
\hat{p}_{i}^{s} = \frac{s_{i,j}^{split}}{\sum s_{i}^{split}}
\end{equation}

Similarly, the Nuclearity Prediction Loss and the Relation Prediction Loss are to maximize the probability of correct nuclear position and discourse relation for each correct split point determined by the decoder respectively.
Since the convergence speed of these three parts is different during the training process, we take the combined one (Eq.~\ref{eq_combine_fun}) as the final loss function and adjust the parameters on the development set.

\begin{equation}\label{eq_combine_fun}
L = \alpha_{s} L_{s} + \alpha_{n} L_{n} + \alpha_{r} L_{r}
\end{equation}

\section{Experimentation}
In this section, we systematically evaluate our top-down text-level discourse parser.

\subsection{Experimental Setting}

\subsubsection{Datasets}
In this paper, we use both the English RST Discourse Treebank (RST-DT)~\cite{carlson2001discourse} and the Chinese Connective-driven Discourse TreeBank (CDTB)~\cite{D14-1224} as the benchmark data sets.

In an RST-style discourse tree, the leaf nodes are non-overlapping text spans called elementary discourse units (EDUs), and internal nodes are the concatenation of continuous EDUs. Adjacent nodes are related through particular discourse relations to form a discourse subtree, which is related to other adjacent nodes in the tree structure. In this way, the hierarchical tree structure is established. The English RST-DT corpus is annotated under the framework of RST. Each document is represented as one DT. It consists of 385 documents (347 for training and 38 for testing) from the Wall Street Journal. We randomly select 34 documents from the training set as our development set.

The Chinese CDTB corpus is motivated by taking both advantages of the English RST-DT corpus (e.g. the tree structure, the nuclearity representation) and the PDTB corpus (e.g., the connective-driven predict-argument structure)~\cite{L08-1093}.
In the Chinese CDTB corpus, each paragraph is marked as a Connective-driven Discourse Tree (CDT), where its leaf nodes are EDUs, its intermediate nodes represent (insertable) connectives (i.e., discourse relations), and EDUs connected by connectives can be combined into higher level discourse units. Currently, the Chinese CDTB corpus consists of 500 newswire articles, which are further divided into 2336 paragraphs with a CDT representation for one paragraph and 10650 EDUs in total.
We divide the corpus into three parts, i.e., 425 training documents containing 2002 discourse trees and 6967 discourse relations,  25 development documents containing 105 discourse trees and 396 discourse relations, 50 test documents containing 229 discourse trees and 993 discourse relations.

\begin{table}
\begin{small}
\begin{center}
\begin{tabular}{rcc}
\toprule
  Parameter                         &English        &Chinese\\ \hline
 POS Embedding                     &30             &30\\
  EDU Encoder BiGRU                 &256            &256\\
  Encoder BiGRU                     &256            &256\\
  Decoder GRU                       &512            &512\\
  bi-directional GRU                 &256            &256\\
  uni-directional GRU   &512            &512\\
  Dropout                           &0.2           &0.33\\
  Split Point Biaffine Attention MLP    &64         &64\\
  Nuclear Biaffine Attention MLP        &64         &32\\
  Relation Biaffine Attention MLP       &64        &128\\
  Epoch                             &20             &20\\
  Batch Size                        &10             &64\\
  Learning Rate                     &0.001          &0.001\\
  $\alpha_{s}$                      &0.3            &0.3\\
  $\alpha_{n}$                      &1.0            &1.0\\
  $\alpha_{r}$                      &1.0            &1.0\\
  \bottomrule
\end{tabular}
\end{center}
\caption{\label{tb_hyper_parameter} Experimental parameter settings.}
\end{small}
\end{table}

\subsubsection{Evaluation Metrics}

To evaluate the parsing performance, we use three standard ways to measure the performance: unlabeled (i.e., hierarchical spans) and labeled (i.e., nuclearity and relation) F-scores.

Same as previous studies, we evaluate our system with gold EDU segmentation and binarize those non-binary subtrees with right-branching. We use the 18 fine-grained relations defined in ~\cite{carlson2001discourse} and the 16 fine-grained relations defined in ~\cite{D14-1224} to evaluate the relation metric for English and Chinese respectively.
In order to avoid the problem that the performance with RST-Parseval evaluation~\cite{marcus2000book} looks unreasonably high, we follow Morey et al.~\shortcite{morey2018dependency}, which adopts the standard Parseval procedure.
For fair comparison, we report micro-averaged $F_1$ scores by default.

\subsubsection{Hyper-parameters}

We use the word embedding representation based on the 300D vectors provided by Glove~\shortcite{pennington2014glove}\footnote{Impact of other pre-trained word embedding is limited. For example, ELMo can improve the full-score about 0.6\%.} and Qiu~\shortcite{qiu2018revisiting} for English and Chinese respectively, and do not update the weights of these vectors during training, while the POS embedding uses the random initialization method and is optimized with our model. We fine-tune the hyper-parameters on the development set as shown in Table~\ref{tb_hyper_parameter}. 

\begin{table}
\begin{small}
\begin{center}
\begin{tabular}{p{0.34cm} c p{0.4cm} p{0.4cm} p{0.4cm} p{0.5cm}}
\toprule
~& Systems & Bare & Nuc & Rel & Full  \\
\hline
\multirow{6}{*}{EN}
&Top-down(Ours)  & 67.2  &55.5  &45.3  &44.3 \\
&Ji\&Eisenstein\shortcite{ji2014representation}$^{+}$  & 64.1  & 54.2  & \textbf{46.8} & \textbf{46.3} \\
&Feng\&Hirst\shortcite{feng2014linear}$^{+}$  &\textbf{68.6}  &\textbf{55.9}  &45.8  &44.6 \\
&Li et al.\shortcite{li2016discourse}$^{+}$  &64.5 &54.0 &38.1 &36.6  \\
&Braud et al.\shortcite{braud2016multi}  &59.5 &47.2 &34.7 &34.3  \\
&Braud et al.\shortcite{braud2017cross}$^{*}$  &62.7 &54.5 &45.5 &45.1  \\
\hline
\multirow{2}{*}{CN}
&Top-down(Ours) &\textbf{85.2}  &\textbf{57.3}   &\textbf{53.3}   &45.7   \\
&Sun\&Kong\shortcite{Sun2018}(Dup)  &84.8   &55.8   &52.1     &\textbf{47.7} \\
\bottomrule
\end{tabular}
\end{center}
\caption{\label{tb_Overall_results}Performance Comparison.(Bare, bare DRS generation. Nuc, nuclearity determination. Rel, rhetorical relation classification. Full, full discourse parsing. The sign $^{+}$ means the systems with additional hand-crafted features including syntactic, contextual and so on, $^{*}$ means with additional cross-lingual features.)}
\end{small}
\end{table}

\subsection{Experimental Results}

\subsubsection{Overall Performance}

First, Table~\ref{tb_Overall_results} compares the detailed performance of our top-down discourse parser with the state-of-the-art on gold standard EDUs.

For English RST-style text-level discourse parsing, we evaluate our top-down discourse parser on the RST-DT corpus and compare our model with five state-of-the-art systems as mentioned in Morey~\shortcite{morey2018dependency} using the same evaluation metrics.\footnote{We evaluate the discourse parsers proposed by Lin et al.~\shortcite{lin-etal-2019-unified} and Liu et al.~\shortcite{liu-etal-2019-hierarchical} in text-level discourse parsing. However, their achieved performances are much lower than the state-of-the-art systems. The main reason is that their proposed encoders are tailored to small text spans in sentence-level discourse parsing and are not suitable for large text spans in text-level discourse parsing. In following experiments, we no longer compare our system with them.}

\begin{itemize}[leftmargin=*]
  \item Ji and Eisenstein~\shortcite{ji2014representation}, a shift-reduce parser that learns the representation of discourse units and trains an SVM classifier jointly with a lot of hand-crafted features.
  \item Feng and Hirst~\shortcite{feng2014linear}, a two stage greedy parser with linear-chain CRF models.
  \item Li et al.~\shortcite{li2016discourse}, an attention-based hierarchical model along with hand-crafted features.
  \item Braud et al.~\shortcite{braud2016multi}, a sequence-to-sequence parser that is heuristically constrained to build trees with a hierarchical neural model.
  \item Braud et al.~\shortcite{braud2017cross}, a transition-based neural model with a lot of cross-lingual features.
\end{itemize}

For Chinese CDT-style text-level discourse parsing, there are much fewer studies. Sun and Kong~\shortcite{Sun2018} propose a complete transition-based Chinese discourse structure generation framework. However, they only concerned tree structure generation and did not consider discourse relation classification. In fact, just as noted in Wang et al.~\shortcite{wang2017two}, a transition-based model is more appropriate for parsing the bare discourse tree structure due to the data sparsity problem. In addition, since relation classification can benefit from the bare tree structure, a two stage parsing strategy can normally achieve better performance. In comparison, with the support of local contextual information of split points and global high-level discourse structure information, our top-down architecture is able to identify the discourse structure and discourse relations jointly. For fair comparison, we duplicate the approach proposed by Sun and Kong~\shortcite{Sun2018}, and evaluate it under the same experimental settings~\footnote{Sun and Kong~\shortcite{Sun2018} reported their performance using macro-averaged $F_1$ scores. In fact, it increases the weight of shorter documents. For Chinese CDTB, each paragraph is represented as a CDT. Statistics on the distribution of CDT heights shows that, one CDT contains about 4.5 EDUs on average, with the average height about 3.42. In this paper, we report the performance using micro-averaged $F_1$ scores. Furthermore, to gain detailed comparison between bottom-up and top-down approaches, we also report the performance of relation classification and full discourse parsing.}. We call this system as the duplicated system (denoted as ``Dup'').
Table~\ref{tb_Overall_results} shows that,
\begin{table}
\begin{small}
\begin{center}
\begin{tabular}{ccccc}
\toprule
language  & Bare & Nuclearity & Relation & Full \\\hline
EN  & 62.3  &50.1   &40.7   &39.6\\
CN  & 80.2  &53.2   &48.5   &41.7\\
\bottomrule
\end{tabular}
\end{center}
\caption{\label{tb_end2end_performance}Performance under a full automatic setting.}
\end{small}
\end{table}

\begin{table*}[ht]
\begin{small}
\begin{center}
\begin{tabular}{c|c|cc|cc|cc|cc}
\toprule
~~  &~~ &\multicolumn{2}{c|}{Bare} &\multicolumn{2}{c|}{Nuc}     &\multicolumn{2}{c|}{Rel}   &\multicolumn{2}{c}{Full}\\
Height  &Std &$\downarrow$ &$\uparrow$ &$\downarrow$ &$\uparrow$    &$\downarrow$ &$\uparrow$   &$\downarrow$ &$\uparrow$   \\\hline
1	&385	&339	&321		&251&221	&233&215&213&200     \\
2	&220	&183	&184		&117&115	&116&111&94	&101     \\
3	&139	&119	&122		&71	&82		&71	&73	&59	&71      \\
4	&88	    &75	    & 78		&52	&58		&44	&42	&39	&40      \\
5	&44	    &34	    & 37		&17	&21		&16	&21	&10	&16      \\
6	&26	    &18	    & 21		&13	&13		&6	&9	&6	&9      \\
7	&18	    &16	    & 18		&7	&8		&6	&9	&2	&5      \\
$>=8$	&13	    &11	    &10		&0	&0		&0	&0	&0	&0      \\   \hline
Overall	&933   &795	&791	&535 &521  &497 &486   &426    &445 \\
\bottomrule
\end{tabular}
\end{center}
\caption{\label{tb_different_level}Performance over different DT levels. (``$\downarrow$''- Top down approach, ``$\uparrow$''- Bottom up approach)}
\end{small}
\end{table*}

\begin{itemize}[leftmargin=*]
  \item For English, our top-down system achieves comparable performance with the state-of-the-art systems. It is worthwhile to note that, we focus on the effectiveness of our proposed top-down architecture in this paper. The performance of our top-down system is achieved without any other additional features, while other systems employ various additional features. For example, both Ji and Eisenstein~\shortcite{ji2014representation} and Feng and Hirst~\shortcite{feng2014linear} employed many kinds of additional hand-crafted features including syntactic, contextual and so on, while Braud et al.~\shortcite{braud2017cross} resort to additional cross-lingual features and achieve the gain of 3.2, 7.3, 10.8 and 10.8 on the four evaluation metrics respectively in comparison with Braud et al.~\shortcite{braud2016multi}. This indicates the great preference of top-down over bottom-up text-level DRS parsing. This also suggests the great potential of additional carefully designed features, which are worth exploring in the future work.
  \item For Chinese, our top-down text-level DRS parser significantly outperforms Sun and Kong~\shortcite{Sun2018} on bare DRS generation, nuclearity determination and relation classification with all p-values smaller than 0.01 on significate testing. However, we find that our top-down approach achieves relatively poor performance on Full discourse parsing. This maybe due to the effectiveness of the joint learning framework as employed in Sun and Kong~\shortcite{Sun2018}. Traditional shift-reduce approaches cast the parsing task as a triple (i.e., shift/reduce action, nuclearity and relation type) identification task, and learn/predict the triple simultaneously, while our top-down approach divides the discourse parsing task into three independent sub-tasks, i.e., split point ranking, nuclearity determination and relation classification, and optimize our discourse parsing model only using the Negative Log Likelihood Loss. This also applies to the English discourse parser discussed above.
  \item Comparing the results for English and Chinese, Chinese text-level discourse parsing looks better on all performance metrics. This maybe due to the difference between annotation strategies. In English RST-DT corpus, each document is represented as one DT, while in Chinese CDTB, each paragraph is represented as a CDT.
      As a result, the CDTs generally contain fewer EDUs and are relatively short in height.
\end{itemize}

\subsubsection{End-to-end Performance}

Next, Table~\ref{tb_end2end_performance} shows the performance of the end-to-end text-level discourse parser under a full automatic setting. Here, we use the two EDU detectors proposed by Li et al.~\shortcite{li2018segbot} and Li et al.~\shortcite{LiYC2013}
to achieve the auto EDUs for English and Chinese respectively,
and the berkeley parser\footnote{https://github.com/slavpetrov/berkeleyparser} to achieve automatic parse trees.
From the results shown in Table~\ref{tb_end2end_performance} we can find that, in comparison with the overall performance using gold standard EDUs shown in Table ~\ref{tb_Overall_results}, there is a significant performance reduction on all the indicators. This indicates the heavy impact of EDU segmentation.

\subsubsection{Detailed Analysis}
Finally, we take Chinese as an example for a detailed comparative analysis. We duplicate the approach proposed by Sun and Kong~\shortcite{Sun2018} and take this duplicated system as the representative of the bottom-up approach.

Table~\ref{tb_different_level} first compares the results over different DT levels with the gold standard numbers and the correctly identified numbers. It should be noted that, correctly determined nuclearity means both the bare tree node and its nuclearity are correctly recognized. Correctly determined relation means both the bare node and its relation are correctly recognized, and full means all three aspects are correctly recognized. From the results we can find that,
in comparison with the bottom-up approach, the top-down approach can achieve better performance on Bare, Nuc and Rel metrics, while for Full-metric, the performance reduces slightly. Just as noted above, this is due to the difference between the joint learning frameworks behind these two approaches.
Among three aspects, the improvement of nuclearity is most, and bare tree structure is weakest. At each level, the performance of these two approaches varies. This suggests that the bidirectional architecture may be an important direction in the future work.

Since the improvement of nuclearity is significant, we then list the detailed results of these two approaches over different nuclearity categories. Table~\ref{tb_Nuclearity_performance} shows that our top-down approach can determine the ``NS'' and ``SN'' much better than the bottom-up approach. This is consistent with human perception.

\begin{table}
\begin{small}
\begin{center}
\begin{tabular}{cccc}
\toprule
    Approach   &NN     &NS     &SN   \\      \hline
    $\downarrow$     &67.0   &42.2   &33.7 \\
    $\uparrow$     &67.6   &35.4   &24.5   \\
\bottomrule
\end{tabular}
\end{center}
\caption{\label{tb_Nuclearity_performance}Performance on nuclearity determination.}
\end{small}
\end{table}

\begin{table}
\begin{small}
\begin{center}
\begin{tabular}{c|c|ccc}
\toprule
~~  &EDU Num    &Bare	&Nuc	&Rel \\  \hline
\multirow{6}{*}{$\uparrow$}
&1--5	    &94.8	&57.9	&52.0   \\
&6--10    &87.0   &60.7	&58.6	\\
&11--15   &78.0   &50.1	&45.4	\\
&16--20   &56.2	&25.0   &25.0   \\
&21--25   &68.9	&47.0   &42.4	\\
&26--30   &65.4	&26.9	&11.5	\\  \hline
\multirow{6}{*}{$\downarrow$}
&1--5	    &97.0   &67.1	&56.6	\\
&6--10    &86.0   &57.3	&59.9	\\
&11--15   &75.2	&50.3	&41.4	\\
&16--20   &56.2	&25.0   &25.0   \\
&21--25   &76.6	&57.7	&40.8	\\
&26--30   &69.2	&42.3	&19.2	\\
\bottomrule
\end{tabular}
\end{center}
\caption{\label{tb_different_edunum}Performance over different EDU numbers.}
\end{small}
\end{table}

We finally divide the DTs into six groups by EDU number and evaluate the two approaches over different groups. Table~\ref{tb_different_edunum} shows the results. We can find that, our top-down approach achieves better performance on the first, fifth and sixth sets (i.e., the EDU number is 1--5, 21-25 and 26-30 respectively). This suggests that the proposed top-down approach may be more suitable for both end of DTs with others comparable.

\section{Conclusion}
In this paper, we propose a top-down neural architecture to text-level discourse parsing.
In particular, we cast the discourse parsing task as an EDU split point ranking task, where a split point is classified to different levels according to its rank, and the EDUs associated with the split point are arranged accordingly. In this way, we can determine the complete discourse rhetorical structure as a hierarchical tree structure.
Specifically, after encoding the EDUs and EDU split points, an encoder-decoder with an internal stack is employed to generate discourse tree recursively.
Experimentation on the English RST-DT corpus and the Chinese CDTB corpus shows the great effectiveness of our proposed approach. In the future work, we will focus on more effective discourse parsing with additional carefully designed features and joint learning with EDU segmentation.

\section*{Acknowledgements}
The authors would like to thank the anonymous reviewers for the helpful comments.
We are greatly grateful to Cheng Sun for his inspiring ideas and preliminary work.
This work is supported by Artificial Intelligence Emergency Project 61751206 under the National Natural Science Foundation of China, Project 61876118 under the National Natural Science Foundation of China and the Priority Academic Program Development of Jiangsu Higher Education Institutions.

\bibliography{acl2021}
\bibliographystyle{acl_natbib}

\end{CJK}
\end{document}